\documentclass[10pt,conference]{IEEEtran}

\usepackage{cite}
\usepackage{amsmath,graphicx,amsfonts,amsthm,amssymb}
\usepackage{graphicx,color}
\usepackage{textcomp}
\usepackage{xcolor}
\usepackage{hyperref}
\hypersetup{hidelinks=true}
\usepackage{diagbox}
\usepackage{multirow}
\usepackage{algpseudocode,algorithm}
\def\BibTeX{{\rm B\kern-.05em{\sc i\kern-.025em b}\kern-.08em
    T\kern-.1667em\lower.7ex\hbox{E}\kern-.125emX}}
\AtBeginDocument{\definecolor{tmlcncolor}{cmyk}{0.93,0.59,0.15,0.02}\definecolor{NavyBlue}{RGB}{0,86,125}}

\usepackage{bbold}
\usepackage{hyperref}
\usepackage{wrapfig}
\usepackage{microtype}
\usepackage{xfrac}
\usepackage{lipsum}
\usepackage{float}
\usepackage{stfloats}
\usepackage{orcidlink}
\usepackage{mathtools}
\usepackage[group-separator={,}]{siunitx}
\usepackage{subcaption}
\usepackage{booktabs}
\usepackage[capitalise]{cleveref}

\IEEEoverridecommandlockouts

\DeclareMathOperator{\chol}{chol}
\newcommand{\V}[1]{{\boldsymbol{\mathbf{#1}}}}

\begin{document}
\title{A Gaussian Process-based Streaming Algorithm for Prediction of Time Series With Regimes and Outliers\thanks{This work was supported by the National Science Foundation under Award 2212506.
}}

\author{\IEEEauthorblockN{Daniel Waxman}
\IEEEauthorblockA{Department of Electrical and Computer Engineering\\
Stony Brook University\\
Stony Brook, New York, USA\\
Email: daniel.waxman@stonybrook.edu}
\and
\IEEEauthorblockN{Petar M. Djuri\'c}
\IEEEauthorblockA{Department of Electrical and Computer Engineering\\
Stony Brook University\\
Stony Brook, New York, USA\\
Email: petar.djuric@stonybrook.edu}}

\maketitle

\begin{abstract}
Online prediction of time series under regime switching is a widely studied problem in the literature, with many celebrated approaches. Using the non-parametric flexibility of Gaussian processes, the recently proposed \textsc{Intel} algorithm provides a product of experts approach to online prediction of time series under possible regime switching, including the special case of outliers. This is achieved by adaptively combining several candidate models, each reporting their predictive distribution at time $t$. However, the \textsc{Intel} algorithm uses a finite context window approximation to the predictive distribution, the computation of which scales cubically with the maximum lag, or otherwise scales quartically with exact predictive distributions. We introduce \textsc{Lintel}, which uses the \emph{exact} filtering distribution at time $t$ with \emph{constant-time} updates, making the time complexity of the streaming algorithm optimal. We additionally note that the weighting mechanism of \textsc{Intel} is better suited to a mixture of experts approach, 
and propose a fusion policy based on arithmetic averaging for \textsc{Lintel}. We show experimentally that our proposed approach is over five times faster than \textsc{Intel} under reasonable settings with better quality predictions.
\end{abstract}

\IEEEpeerreviewmaketitle

\section{Introduction}

Online prediction for time series is an important problem in many machine learning and signal processing applications. It has accordingly received extensive attention from the perspectives of kernel learning \cite{richard2009online}, recurrent neural networks \cite{guo2016robust}, and Gaussian processes (GPs) \cite{liu2020sequential}, among others. The problem is especially interesting when outliers or regime switches are present in the data, as outliers can contaminate a model's predictions if care is not taken to avoid this.

This paper focuses on improving the INstant TEmporal structure Learning (\textsc{Intel}) algorithm \cite{liu2020sequential}, a recently proposed method that models time series as Gaussian processes, and accounts for outliers and regime switches by dynamically adjusting the current data used for inference. This is accomplished by prespecifying several ``candidate models'' and using a (generalized) product of experts approach for fusing estimates. The weights for each model are also dynamically adjusted in a modified version of Bayesian model averaging (BMA) -- an approach that has recently proved successful in several other Bayesian online learning problems \cite{lu2022incremental,waxman2024doebe}.

One drawback to the GP approach of \textsc{Intel} is the need to recompute the predictive distributions of each candidate GP at every time step. Using the typical kernel approach, the computation of the predictive distribution at time $t$ scales with $\mathcal{O}(t^3)$ and quickly becomes prohibitively expensive. As a result, \textsc{Intel} approximates the predictive distribution, only using the previous $\tau$ data points instead. If $\tau$ is small, the algorithm will be tractable, but as $\tau$ grows smaller, the approximation becomes more severe.

Instead, we propose to use Markovian GPs and their corresponding state-space representation \cite{sarkka2013spatiotemporal}. Markovian GPs admit a representation as a linear time-invariant stochastic differential equation, which allows for online, exact predictive distributions via Kalman filtering. Accordingly, our algorithm \textsc{Lintel} (short for Linear \textsc{Intel}) provides constant time-complexity updates using the exact predictive distribution. Using the state-space representation of GPs, we also draw connections between the \textsc{Intel} algorithm and the $3\sigma$-rejection Kalman filter \cite[Ch. 7]{zoubir2018robust}. We additionally make our code available open-source at \url{https://www.github.com/DanWaxman/Lintel}.

The rest of this paper is structured as follows: in \cref{sec:gpr}, we provide background in GPs, which is used in \cref{sec:intel_algorithm}, where we discuss the \textsc{Intel} algorithm. In \cref{sec:lin_time_gpr}, we extend our discussion on GPs to filtering for inference with linear time complexity. This is used in \cref{sec:lintel_algo}, where the \textsc{Lintel} algorithm is introduced. We perform a series of experiments on synthetic and real data in \cref{sec:experiments}, followed by brief discussion and conclusions in \cref{sec:conclusions}.

\section{Gaussian Process Regression} \label{sec:gpr}
GPs are a powerful tool in Bayesian machine learning, being non-parametric, flexible, and theoretically tractable \cite{rasmussen_and_williams}. In this section, we provide a brief overview of their use in GP regression using kernels (\cref{sec:gpr_with_kernels}), and some computational drawbacks associated with the kernel approach (\cref{sec:gpr_drawbacks}). 

\subsection{Gaussian Process Regression with Kernels} \label{sec:gpr_with_kernels}
GPs are stochastic processes $\{f(t)\}_{t \in \mathcal{T}}$ defined by the property that for any finite index set $\{ t_1, \dots, t_{N_\text{data}} \}$, the random variables $\{f(t_1), \dots, f(t_{N_\text{data}})\}$ are jointly Gaussian distributed. In the signal processing and machine learning communities, GPs are typically presented in a functional space with a kernel formulation, whereby the covariance between two input points $\V{x}, \V{x'}$ is specified by a kernel function $\kappa(\V{x}, \V{x'})$. By additionally specifying a mean function $\mu(t)$, the GP is fully determined, and written as $f \sim \mathcal{GP}(\mu, \kappa)$.

In GP regression, we place a GP prior on an unknown function $f$ and specify a likelihood for $y | t, f$ to obtain a posterior GP. The most common likelihood is also Gaussian, in which case the data-generating process is described by
\begin{align*} 
    f &\sim \mathcal{GP}(\mu, \kappa), \\
    y &= f(t) + \varepsilon,
\end{align*}
where the observation noise $\varepsilon$ is i.i.d. Gaussian with variance $\sigma_n^2$. When Gaussian likelihoods are used, the model is conjugate, allowing for analytic posterior computations. In particular, given training data $\mathcal{D} = \{ (t_n, y_n) \}_{n=1}^{N_\text{data}}$, the predictive distribution of the posterior GP at input $t_*$ is given by 
\begin{align}
    f(t_*) | \mathcal{D} &\sim \mathcal{N}(m_*, s^2_*), \label{eq:gp_pred_dist} \\
    m_* &= \mu(t) + \V{k}_{*\odot} \left(\V{K}_{\odot\odot} + \sigma_n^2 \mathbb{1}\right)^{-1} \V{\delta}, \label{eq:gp_pred_mean} \\
    s^2_* &= K_{**} - \V{k}_{*\odot} \left(\V{K}_{\odot\odot} + \sigma_n^2 \mathbb{1}\right)^{-1} \V{k}_{\odot*}, \label{eq:gp_pred_var}
\end{align}
where $[\V{K}_{\odot\odot}]_{ij} \triangleq \kappa(t_i, t_j)$ is the cross-covariance matrix of the training data, $\V{k}_{*\odot} = \V{k}_{\odot*}^\top$ is a row vector whose entries are $\kappa(t_i, t_*)$, $k_{**} = \kappa(t_*, t_*)$ is the variance at the test point, and the mean function is used for the row vector $\V{\delta}$ whose entries are $y_n - \mu(t_n).$

A typical example of kernel function which is used in this work is the Mat\'ern-$5/2$ kernel, which is given by
\begin{equation}
    k_{\text{Mat-}5/2}(t, t') = \sigma_f^2 \left(1 + \frac{\sqrt{5}}{3\ell} + \frac{5}{3\ell^2} \right) \exp\left(-\frac{\sqrt{5} r}{\ell}\right),
\end{equation}
where $r = \lvert t - t' \rvert$ and the process scale $\sigma_f$ and length scale $\ell$ are treated as hyperparameters. GPs with kernels which only depend on $r$ are known as stationary. The hyperparameters, which we will denote as $\V\psi$, are often trained by maximizing the evidence.

\subsection{Drawbacks of the Kernel Approach} \label{sec:gpr_drawbacks}
While the predictive quantities given in \cref{eq:gp_pred_dist,eq:gp_pred_mean,eq:gp_pred_var} are analytically tractable, their expressions involve the inversion of an $N_\text{data} \times N_\text{data}$ matrix. In practice, the memory cost (and some of the computational cost) of a full matrix inversion can be avoided by instead solving the corresponding linear systems using Cholesky decompositions (see, e.g., \cite[App. A]{rasmussen_and_williams}). However, the computational complexity still scales cubically with $N_\text{data}$, making exact GP inference prohibitively costly for large datasets. This can be particularly problematic in the context of long time series, where $N_\text{data}$ can be several thousands of points long.

Another drawback of using kernels for GP regression is difficulties with updating the training set. In particular, in an online setting, we are often interested in the predictive distribution $p(y_n | y_{1:n-1})$; using kernels, it is generally difficult to directly use $p(y_n | y_{1:n-1})$ to calculate $p(y_{n+1} | y_{1:n})$ without recomputing some part of the Cholesky decomposition of $\V{K}_{\odot\odot}$, and one must resort to approximations for efficient (i.e., bounded) online updates \cite{ranganathan2010online,huber2014recursive}. Note that in the remainder of the paper, the index $n$ corresponds to the time instant $t_n$.

\section{The \textsc{Intel} Algorithm} \label{sec:intel_algorithm}
The \textsc{Intel} algorithm \cite{liu2020sequential} uses several candidate GPs $f_1, \dots, f_K$ that are fused to perform online prediction, allowing for the presence of outliers or the possibility of regime switches. We first describe the weighting and fusion mechanism of \textsc{Intel} (\cref{sec:intel_fusion}), and subsequently its outlier and change point detection mechanism (\cref{sec:intel_cpd}). We conclude with some notes about initialization and various approximations used in the algorithm (\cref{sec:intel_approx}).

\subsection{Weighting and Fusing} \label{sec:intel_fusion}
The \textsc{Intel} algorithm fuses the predictions of candidate GPs $f_1, \dots, f_K$ with predictive densities $p_1(y_{n+1} | \mathcal{D}_n), \dots, p_K(y_{n+1} | \mathcal{D}_n)$ using a weighted product of experts framework \cite{hinton2002training}, where $\mathcal{D}_n$ includes all data at the current time used for prediction --- we will elaborate more on its contents in \cref{sec:intel_cpd}. Each GP is assumed to have its own hyperparameters $\V \psi_k$, but share a common (constant) mean function $\mu(\cdot) = C$. We first describe the weights, and then the fusion rule, since the former is required for the latter.

The weighting in \textsc{Intel} is based on a modification to BMA where a forgetting factor $0 \leq \alpha \leq 1$ is used, leading to a recursive update to the filtered weight
\begin{equation} \label{eq:w_tilde}
    \Tilde{w}_{k, n+1} \propto w_{k, n}^\alpha,
\end{equation}
which are normalized to sum to unity and used for prediction. After $y_{t+1}$ is observed, the weights are updated using the likelihood $p_k(y_{n+1} | \mathcal{D}_n)$ according to the rule
\begin{equation} \label{eq:w}
    w_{k, n+1} \propto \Tilde{w}_{k, n+1} p_k(y_{n+1} | \mathcal{D}_n),
\end{equation}
which are once again normalized to sum to unity.

Once the weights are determined, the predictive densities of each candidate model are geometrically averaged. In particular, a weighted product of experts is proposed, where 
\begin{equation*}
    p(y_{n+1} | \mathcal{D}_n) \propto \prod_{k=1}^K p_k(y_{n+1} | \mathcal{D}_n)^{\Tilde{w}_k}.
\end{equation*}
In the GP literature, such a model is known as a generalized product of experts, and was introduced in \cite{cao2014generalized}. Since each $p_k(\cdot | \cdot)$ is Gaussian, $p(y_{n+1} | \mathcal{D}_n)$ is also Gaussian, with mean and variance
\begin{align}
    m_{n+1} &= \frac{\sum_{k=1}^K m_{k, n+1} / \Tilde{w}_{k, n+1} \sigma^{-2}_{k, n+1}}{\sum_{k=1}^K \Tilde{w}_{k, n+1} \sigma^{-2}_{k, t+1}}, \label{eq:gpoe_mean} \\
    s^2_{t+1} &= \left( \sum_{k=1}^K \Tilde{w}_{k, n+1} \sigma^{-2}_{k, t+1} \right)^{-1}. \label{eq:gpoe_var}
\end{align}

For a fixed set of weights that sum to unity, arithmetic and geometric averaging can be understood to minimize a weighted sum of Kullback-Liebler divergences in opposite directions \cite{li2019second}. Typical methods to determine weights include relative entropy with respect to the prior \cite{cao2014generalized}, optimizing functionals of the covariance matrix \cite{hurley2002information}, and Bayesian learning \cite{carvalho2023bayesian}, which can also be input-dependent \cite{ajirak2023fusion}. The selection of weights is important, as the relative predictive performance of arithmetic and geometric fusion depends on the choice of weights \cite{li2019second}.

\subsection{Outlier and Change Point Detection} \label{sec:intel_cpd}
Using the fused estimate, a $3\sigma$ credible interval is constructed. If the data point $y_{n+1}$ is within the $3\sigma$ credible interval $[ m_{n+1} - 3s_{n+1}, m_{n+1} + 3s_{n+1}]$, it is then added to the data $\mathcal{D}_n$. Otherwise, it is declared an outlier and added to the potential changepoint bucket (PCB), denoted by $\mathcal{D}'_n$. If the PCB reaches a predetermined length $N_\text{PCB}$, a regime-switch is announced; in this case, the data is reset with $\mathcal{D}_n \gets \mathcal{D}_n'$, and $\mu(\cdot)$ is set to the constant function with mean given by $\frac{1}{N_\text{PCB}} \sum_{y \in \mathcal{D}_n'} y$. After a datum that is not an outlier is observed, the PCB is reset, $\mathcal{D}_n' \gets \{\}$.

Note that this detection mechanism for outliers and regime switches is not perfect; for example, a regime switch might \emph{reduce} the variance of the observation noise, in which case all points would lay well within $3\sigma$. However, in this case, the weighting and fusion mechanism will rapidly adapt to the ``correct'' model. In this sense, the PCB is not the only mechanism for regime switches, but serves as a rapid detection mechanism when extreme and sudden changes occur.

\subsection{Initialization and Approximations} \label{sec:intel_approx}
As previously mentioned in \cref{sec:gpr_drawbacks}, the computation of $p(y_{n+1} | \mathcal{D}_n)$ requires $\mathcal{O}(n^3)$ time, which is prohibitive as $n$ grows large. As a result, the authors of $\textsc{Intel}$ propose using a finite context window of length $\tau$, using $p(y_{n+1} | \Tilde{\mathcal{D}}_n)$ instead of $p(y_{n+1} | \mathcal{D}_n)$ where $\Tilde{\mathcal{D}}_n$ is the intersection of $\mathcal{D}_n$ and the finite context window of length $\tau$, i.e.,
\begin{equation*}
    \Tilde{\mathcal{D}}_n = \mathcal{D}_n \cap \{ (t_{n'}, y_{n'})\}_{n'=n-\tau+1}^{n}.
\end{equation*} The GP is then recomputed at every timestep using the last $\tau$ observations, resulting in $\mathcal{O}(\tau^3)$ updates. 

To create a list of candidate models, a small subset of data $\mathcal{D}_0$ is set aside, and the hyperparameters are selected by evidence maximization. The authors of \textsc{Intel} then advocate for creating several candidate models using prior knowledge about the system; for example, in a CPU utilization data, they choose a candidate model such that the process scale is $1/5$th that of the evidence maximization estimate, based on prior knowledge that volatility might shrink.

Two general remarks are in order:
\paragraph*{Remark 1} Note that the Cholesky decomposition $L = \chol(\V{K}_{\odot\odot} + \sigma_{\varepsilon}^2 \mathbb{1})$ must be recomputed at each time step because we do not assume that the data $t_1, \dots, t_{N_\text{data}}$ are equally-spaced. 
Even so, an improvement can be made to the \textsc{Intel} algorithm by performing rank-1 updates to $L$, which can be performed in $\mathcal{O}(\tau^2)$ time. While these updates are worthwhile in practice, our proposed method allows for constant-time updates without windowing.

\paragraph*{Remark 2} The typical choice of fusion rule when using BMA-style weighting is an arithmetic average, rather than a geometric average, which comes with several desirable properties in prediction \cite{liu2023robust}. While a mixture of experts approach is not intrinsically superior to a product of experts approach for any fixed weight 
(see, e.g., the discussion in \cite{li2019second}),
there are no theoretical guarantees that the BMA-style weights are optimal for geometric pooling with respect to, for example, predictive log-likelihood. Our approach will therefore adopt the mixture of experts (i.e., arithmetic fusion) approach instead.

Pseudocode for implementing \textsc{Intel} is available in \cref{alg:intel}.

\begin{algorithm}[t]
\caption{\textsc{Intel} \cite{liu2020sequential}}\label{alg:intel}
\begin{algorithmic}[1]
\Require Window Size $\tau$, Maximum PCB Size $N_{\text{PCB}}$, Prior Mean $C$, Forgetting Factor $\alpha$, Mean Update Period $L$, Initial Weights $\V{w}_0$, Kernel Hyperparameters $\V\psi_1, \dots, \V\psi_K$
\Ensure Output mean and variance $m_n, s^2_n$ and outlier flag $\texttt{is\_outlier}_n$ for $n = 1, 2, \dots$
\State $\Tilde{\mathcal{D}} \gets \{ \}$
\State $\mathcal{D}' \gets \{ \}$
\State $t_{\text{last\_mean\_update}} \gets 0$
\For{$n=1, 2, \dots$}
    \State Receive input $t_n$
    \For{$k=1, \dots, K$}
        \State Calculate $m_{k, n+1}, s^2_{k, n+1}$ from \cref{eq:gp_pred_mean,eq:gp_pred_var}
    \EndFor
    \State Calculate $\V{\Tilde{w}}_{n+1}$ from \cref{eq:w_tilde}
    \State Receive output $y_{n}$
    \State Calculate $m_{n+1}, s^2_{n+1}$ from \cref{eq:gpoe_mean,eq:gpoe_var}
    \State $\texttt{is\_outlier} \gets y_{n} \in [ m_{n+1} - 3s_{n+1}, m_{n+1} + 3s_{n+1}]$
    \If{$\neg\texttt{is\_outlier}$}
        \State $\mathcal{D}' \gets \{ \}$
        \State{Add $(t_n, y_n)$ to $\Tilde{\mathcal{D}}$}
        \If{$ t_{\text{last\_mean\_update}} \geq L$}
            \State Update $C$ with the average of $\Tilde{\mathcal{D}}$
            \State $t_{\text{last\_mean\_update}} \gets 0$
        \Else
            \State $t_{\text{last\_mean\_update}} \gets t_{\text{last\_mean\_update}} + 1$
        \EndIf
    \Else
        \State{Add $(t_n, y_n)$ to $\mathcal{D}'$}
        \If{$\lvert \mathcal{D}' \rvert \geq N_\text{PCB}$} \Comment{Declare Changepoint}
            \State $\mathcal{D} \gets \mathcal{D}'$ 
            \State Update $C$ with the average of $\mathcal{D}$
            \State $t_{\text{last\_mean\_update}} \gets 0$
        \EndIf
    \EndIf
\EndFor
\end{algorithmic}
\end{algorithm}

\section{Linear-Time Gaussian Process Regression} \label{sec:lin_time_gpr}
In this section, we introduce the linear-time inference of GP regression, which relies on the representation of GP as a stochastic differential equation and performs inference via Kalman filtering and smoothing \cite{sarkka2013spatiotemporal}. These will form the basis of our proposed method. GPs that fit into this framework are often called \emph{Markovian GPs} or \emph{state-space GPs}. Any GP with with a stationary kernel  (i.e., $\kappa(x, x')$ expressed solely in terms of the difference $r = \lvert x - x' \rvert$) can be approximated arbitrarily well through Markovian GPs \cite{sarkka2013spatiotemporal}. We provide an overview of the state-space representation of Markovian GPs (\cref{sec:markov_gp_ssm}) and their inference using Kalman filtering (\cref{sec:markov_gp_filtering}).

\subsection{Markovian Gaussian Processes} \label{sec:markov_gp_ssm}
The $\mathcal{O}(N_\text{data}^3)$ complexity of inference in GP regression has been a major practical issue, leading to many approximate inference methods over the last 20 years. However, for special choices of the covariance function in scalar GPs, \emph{exact} inference can be performed in $\mathcal{O}(N_\text{data})$ time by viewing GPs through the lens of stochastic differential equations (SDEs) \cite{sarkka2013spatiotemporal}. 

Exact details of SDEs are beyond the scope of this work, but with several technical caveats, we may view these as ordinary differential equations driven by some noise process. The most common noise process is a \emph{Brownian motion} $\V{\beta}(t) \in \mathbb{R}^k$ with diffusion matrix $\V Q$, which is defined by the following three properties \cite[Def. 4.1]{sarkka2019applied}: (1) $\V{\beta}(0)$ is zero, (2) increments are independent for independent time intervals, and (3) $\V\beta(t_1) - \V\beta(t_2) \sim \mathcal{N}(\mathbf{0}, (t_2 - t_1) \V{Q})$. For further details on SDEs, the reader is directed to the excellent texts of S\"arkk\"a \& Solin \cite{sarkka2019applied} and {\O}ksendal \cite{oksendal2013stochastic}.

In this setting, the GP prior is represented as a continuous-discrete linear Gaussian state space:
\begin{align*}
    d{\V\theta} &= \V{F} \V\theta + \V{L} \,d\V{\beta}, \\
    y_n &= \V{h}^\top \V\theta(t_n) + \mu(t_n) + \varepsilon,
\end{align*}
where $\V{\beta}$ is a Brownian motion with diffusion matrix $\V Q$ and $\varepsilon$ is i.i.d. Gaussian noise with variance $\sigma_n^2$. This is discretized to a discrete-time state space model \cite[Ch. 10.6]{sarkka2019applied}:
\begin{align}
    {\V\theta}(t_{n+1}) &= \V A_n {\V\theta}(t_n) + \V{q}_n, \label{eq:markov_gp_trans_eq} \\
    y_n &= \V{h}^\top {\V\theta}(t_n) + \mu(t_n) + \varepsilon_n, \label{eq:markov_gp_obs_eq}
\end{align}
where $\V{A}_n$ and the covariance $\V \Sigma$ of $\V{q}_n$ can be derived from the SDE. In particular, $\V{A}_n$ is related to the continuous transition matrix $\V{F}$ and the time between successive datapoints $\Delta t_n = t_n - t_{n-1}$ by $\exp(\V{F} \Delta t_n)$. The notation $\V\theta(t_{n})$ (as opposed to $\V\theta_n$) reinforces that $\V\theta(t_{n})$ arises from a continuously-indexed stochastic process.
Using the discrete-time state space model, a solution with linear time complexity in $N_\text{data}$ can then be computed using standard Kalman filtering and smoothing. GPs with many different covariance matrices (notably, Mat\'ern covariance matrices) can be represented this way, and even more (e.g., squared exponential kernels) can be approximated similarly --- see \cite[Ch. 11]{sarkka2019applied} for several examples.

\subsection{Inference With Kalman Filtering} \label{sec:markov_gp_filtering}
In the case of online inference, we are principally concerned with filtering solutions. The filtering equations for Markovian GPs are the standard Kalman filter ones, which track the current mean and covariance of $\V{\theta}(t_n)$, denoted as $\V m_n$ and $\V P_n$, respectively. The prediction density $p(\V{\theta}(t_n) | y_{1:n-1})$ is a normal with mean and covariance
\begin{align}
    \V m_n^{-} &= \V A_{n-1} \V m_{n-1}, \\
    \V P_n^{-} &= \V A_{n-1} \V P_{n-1} \V A_{n-1}^\top + \V \Sigma.
\end{align}
The predictive distribution $p(y_n | y_{1:n-1})$ is also normal with mean and covariance
\begin{align}
    m_n &= \V{h}^\top \V{m}_n^{-} + \mu(t_n), \label{eq:markov_gp_predictive_mean} \\
    s^2_n &= \V{h}^\top \V{P}_n^{-1} \V{h} + \sigma_n^2. \label{eq:markov_gp_predictive_var}
\end{align}
The filtering distribution $p(\V{\theta}(t_n) | y_{1:n})$
is then once again normal with mean and covariance
\begin{align}
    \V{m}_n &= \V{m}_n^{-} + \V{k}_n (y_n - m_n), \label{eq:markov_gp_filter_mean} \\
    \V{P}_n &= \V{P}_n^{-} - \V{k}_n \V{S}_n \V{k}_n^\top, \label{eq:markov_gp_filter_covar}
\end{align}
where $\V{k}_n$ is the \emph{Kalman gain} given by
\begin{equation}
    \V{k}_n = \V{P}_n^{-} \V{h} / s_n^2. \label{eq:markov_gp_kalman_gain}
\end{equation}

\section{The \textsc{Lintel} algorithm} \label{sec:lintel_algo}
The form of the \textsc{Intel} algorithm is particularly well-suited for linear time GPR. In fact, using linear GPs with exact filtering, we can even avoid using finite context window approximations to the predictive distribution while maintaining constant-time updates. Because of its strictly linear time complexity with respect to the length of the dataset, we call our algorithm \textsc{Lintel}. In this section, we introduce the \textsc{Lintel} algorithm (\cref{sec:lintel_details}). We then discuss how updates to the mean function should be interpreted (\cref{sec:lintel_mean_update}), and relate the algorithm to previous work in the GP and robust filtering communities (\cref{sec:lintel_related_work}).

\begin{algorithm}[t]
\caption{\textsc{Lintel}}\label{alg:lintel}
\begin{algorithmic}[1]
\Require Maximum PCB Size $N_{\text{PCB}}$, Prior Mean $C$, Forgetting Factor $\alpha$, Mean Update Period $L$, Initial Weights $\V{w}_0$, Kernel Hyperparameters $\V\psi_1, \dots, \V\psi_M$
\Ensure Output mean and variance $m_n, s^2_n$ and outlier flag $\texttt{is\_outlier}_n$ for $n = 1, 2, \dots$
\State $\mathcal{D} \gets \{ \}$
\State $t_\text{last} = t_1$
\State $t_{\text{last\_mean\_update}} \gets 0$
\For{$n=1, 2, \dots$}
    \State Receive input $t_n$
    \For{$k=1, \dots, K$}
        \State Set $\Delta t = t_n - t_\text{last}$ and calculate $\V{A}_n$
        \State Calculate $m_{k, n+1}, s^2_{k, n+1}$ from \cref{eq:markov_gp_predictive_mean,eq:markov_gp_predictive_var}
    \EndFor
    \State Calculate $\V{\Tilde{w}}_{n+1}$ from \cref{eq:w_tilde}
    \State Receive output $y_{n}$
    \State Calculate $m_{n+1}, s^2_{n+1}$ from \cref{eq:moe_mean,eq:moe_var}
    \State $\texttt{is\_outlier} \gets y_{n} \in [ m_{n+1} - 3s_{n+1}, m_{n+1} + 3s_{n+1}]$
    \If{$\neg\texttt{is\_outlier}$}
        \State Calculate filtering distributions from \cref{eq:markov_gp_filter_mean,eq:markov_gp_filter_covar,eq:markov_gp_kalman_gain}
        \State $t_\text{last} \gets t_n$
        \If{$ t_{\text{last\_mean\_update}} \geq L$}
            \State Update $C$ with the average of $\mathcal{D}$
            \State Update $\V{m}_n$ with \cref{eq:mean_update}
            \State $t_{\text{last\_mean\_update}} \gets 0$
        \Else
            \State $t_{\text{last\_mean\_update}} \gets t_{\text{last\_mean\_update}} + 1$
        \EndIf
    \Else
        \State{Add $(t_n, y_n)$ to $\mathcal{D}'$}
        \If{$\lvert \mathcal{D}' \rvert \geq N_\text{PCB}$} \Comment{Declare Changepoint}
            \State $\mathcal{D} \gets \mathcal{D}'$ 
            \State $t_\text{last} \gets t_n$
            \State Update $C$ with the average of $\mathcal{D}$
            \State $t_{\text{last\_mean\_update}} \gets 0$
            \State Perform Kalman filtering on $\mathcal{D}$ to update $\V\theta(t_n)$
        \EndIf
    \EndIf
\EndFor
\end{algorithmic}
\end{algorithm}

\subsection{The \textsc{Lintel} Algorithm} \label{sec:lintel_details}
The \textsc{Lintel} algorithm falls very quickly from the adoption of Markovian GPs in the \textsc{Intel} algorithm, and the basic idea is simple: instead of keeping track of $\mathcal{D}_n$ and recomputing GP posteriors at each time step, we instead keep track of the filtered state $\V{\theta}(t_n)$, which provides constant-time Bayesian updates to the GP predictive. In particular, the weighting mechanism of \cref{sec:intel_fusion} stays the same, as does the basic mechanics of outlier and changepoint detection. We propose only two simple changes:

\begin{enumerate}
    \item Use the filtering equations \cref{eq:markov_gp_filter_mean,eq:markov_gp_filter_covar,eq:markov_gp_kalman_gain} and predictive equations \cref{eq:markov_gp_predictive_mean,eq:markov_gp_predictive_var} to perform online GP inference. Accordingly, we use the entire (correct) predictive distribution of the current regime, $p(y_n | \mathcal{D}_n)$, at no additional cost.
    \item Use arithmetic averaging as the fusion rule, rather than geometric averaging.
\end{enumerate}

For (1), implementation is trivial after the corresponding state-space model is obtained. The only catch is that now we have to ``backtrack'' over the previous $N_\text{PCB}$ points when a changepoint is declared. However, this operation remains linear in $N_\text{PCB}$, unlike the  corresponding operation in the \textsc{Intel} algorithm, which is cubic in $N_\text{PCB}$.

For (2), we in principle obtain a predictive distribution that is a mixture of Gaussians. For simplification, we follow recent work with ensembles of GPs (e.g., \cite{lu2022incremental,waxman2024doebe}),  in recording the minimum mean square error Gaussian estimate $m_{n+1}$ and its variance, $s_{t+1}^2$. They are given by
\begin{align}
    m_{n+1} &= \sum_{k=1}^K \Tilde{w}_{k, n+1} m_{k, n+1} , \label{eq:moe_mean} \\
    s^2_{n+1} &= \sum_{k=1}^K \left( \sigma_{k, n+1}^{2} + (m_{n+1} - m_{k, n+1})^2 \right) \Tilde{w}_{k, n+1}. \label{eq:moe_var}
\end{align}

\subsection{Updating the Mean Function} \label{sec:lintel_mean_update}
Care must be taken to understand how periodically updating the mean is to be interpreted. A first approach would be to directly update $\mu(t_n)$ and use \cref{eq:markov_gp_obs_eq} as the likelihood. However, unlike in \textsc{Intel}, changes in the mean function now correspond to steps in a piecewise-constant mean function in the GP prior. Accordingly, changing the mean from $C$ to $C'$ at time $t_n$ corresponds to a discontinuous ``jump'' of magnitude $\Delta C \triangleq C' - C$ in the prior mean. 

However, in our case, we likely want $p(y_{n+1} | \mathcal{D}_n)$ to remain the same after updating the mean. Fortunately, this is convenient and efficient to implement by considering the augmented state space \cref{eq:markov_gp_trans_eq,eq:markov_gp_obs_eq}. In particular, $p(y_{n+1} | \mathcal{D}_n)$ will be the same after updating the mean if the change in mean and a change in $\V{h}^\top \V{\theta}(t_n)$ cancel each other. Then, we must change the current filtering mean $\V{m}_n$ accordingly.

There are an infinite number of choices when performing the corresponding update; assuming without loss of generality that $\V{h}$ is a vector of zeros and ones, the simplest such update corresponds to reducing all latent function values by an average of $\Delta C$, i.e., 
\begin{equation*}
    \V{m}_n \to \V{m}_n - \frac{\V{h} \Delta C}{\lVert \V{h} \rVert_{\ell_1}}.    
\end{equation*}
However, it seems intuitive to instead update proportional to the current function values. For example, suppose a GP is comprised of a quasiperiodic component and a trend component. In that case, if the current trend is near zero and the quasiperiodic component is very large, the latent function values corresponding to the quasiperiodic function should bear most of the update. The corresponding update, which we use in \textsc{Lintel}, is given by
\begin{equation} \label{eq:mean_update}
    \V{m}_n \to \V{m}_n - \frac{\left(\V{h} \odot \V{m}_n\right)\Delta C}{\V{h}^\top \V{m}_n},
\end{equation}
where $\V{a} \odot \V{b}$ is the Hadamard (elementwise) product.
One may interpret this update as placing a new (informative) GP prior on the space using the updated mean function. From a filtering perspective, it is convenient to think of this update as creating a new model with a well-motivated initial distribution $p(\V{\theta})$.

Pseudocode for implementing \textsc{Lintel} is available in \cref{alg:lintel}. Initial hyperparameters may be determined in the same manner as for \textsc{Intel}, i.e., by maximizing the evidence over some ``pretraining'' points.

\subsection{Related Work} \label{sec:lintel_related_work}
For outlier detection, using a credible interval of the predictive distribution is well-studied. Indeed, the $3\sigma$-rejection Kalman filter is established as a simple way to robustify the Kalman filter \cite[Ch. 7]{zoubir2018robust}, and its applications and limitations have been studied in the applied filtering literature \cite{berman2014outliers}.

In fact, an outlier-rejection Kalman filter was similarly used with Markovian GPs in \cite{bock2022online}. There, GP factor analysis is combined with Markovian GPs to provide outlier-robust GP inference, where outliers are determined online by a threshold to the log-likelihood. Their work, however, does not incorporate the possibility of regime switching, the fusion of several candidate models, or adaptive means. Notably, in \textsc{Lintel}, the credible region is constructed based on the fused estimate and not the individual estimates.

\section{Experiments \& Discussion} \label{sec:experiments}

To assess the performance of \textsc{Lintel}, we first test \textsc{Intel} and \textsc{Lintel} on a synthetic dataset where only outliers are present (\cref{sec:experiments_syn_only_outliers}). This is followed by a similar example where regime switching occurs (\cref{sec:experiments_syn_with_regime}). Finally, we experiment on a real-world datasets (\cref{sec:experiments_real_data}). All code to reproduce experiments is available at \url{https://www.github.com/DanWaxman/Lintel}.

\subsection{Synethetic Data With Outliers} \label{sec:experiments_syn_only_outliers}
\begin{figure*}[th]
    \centering
    \includegraphics[width=0.80\textwidth]{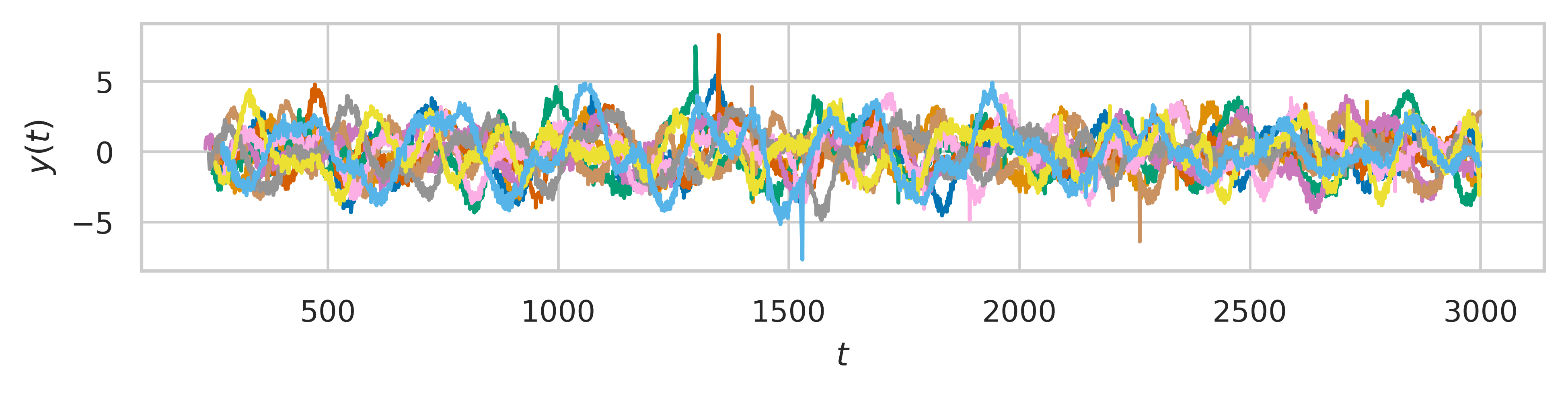}
    \caption{Data used in the synthetic data with outliers experiment, with each color representing a different random seed.}
    \label{fig:experiment_1_data}
\end{figure*}

\begin{figure*}[th]
    \centering
    \includegraphics[width=0.8\textwidth]{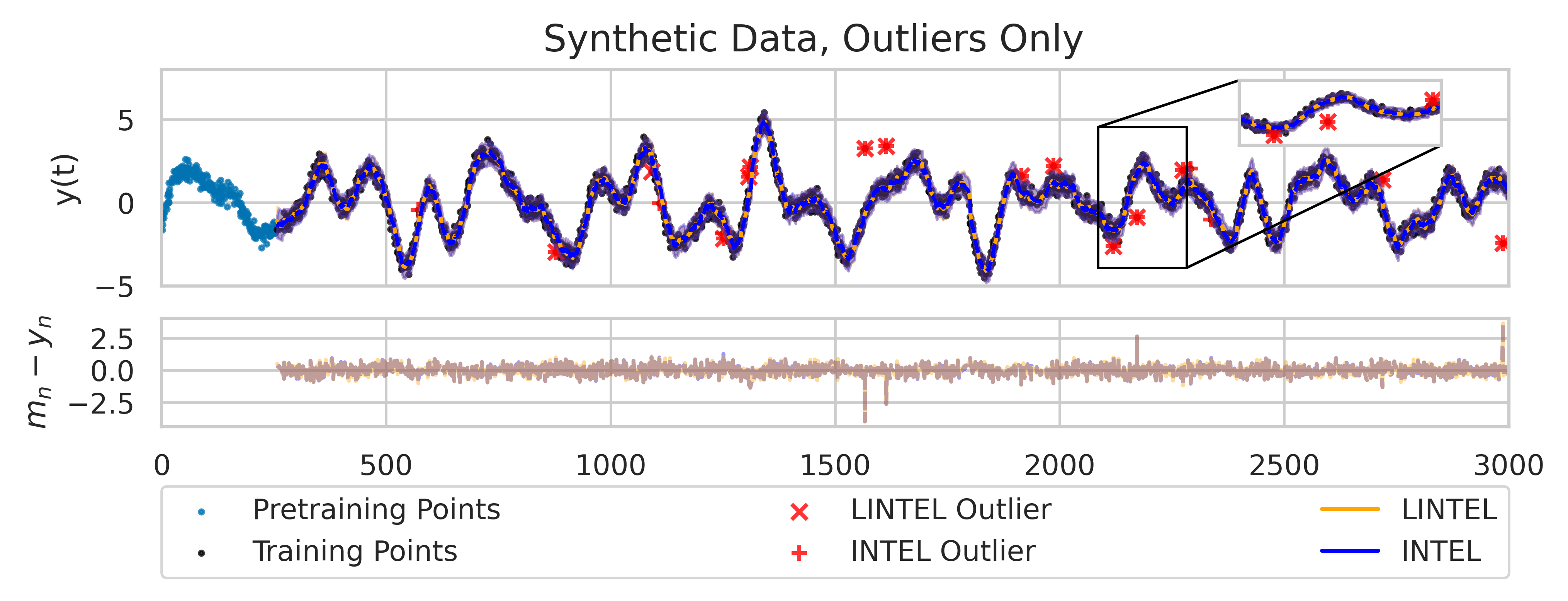}
    \caption{An example of the output for the synthetic data with outliers experiment. {\textbf{Top}:} The outputs of \textsc{Intel} and \textsc{Lintel}, with reported outliers marked. Shaded regions denote two standard deviations. {\textbf{Bottom:}} The difference in predictive mean $m_n$ and the data point $y_n$.}
    \label{fig:experiment_1_example}
\end{figure*}

\begin{table*}[t]
\centering
\caption{Results for Synthetic Data, Outliers Only Experiment. Higher is better for mean predictive-log-likelihood, and lower is better for normalized mean square error and running time. The best result (significant at the $p=0.01$ level according to a Wilcoxon signed rank-sum test) is bolded.}
\begin{tabular}{lcccc}\toprule
&\multicolumn{2}{c}{\textbf{\textsc{Intel}}} & \multicolumn{2}{c}{\textbf{\textsc{Lintel}}} \\\cmidrule(lr){2-3}\cmidrule(lr){4-5} 
& Arithmetic & Geometric & Arithmetic & Geometric \\\midrule
{Mean Predictive Log-Likelihood}     & {$-0.375\pm 0.049$} & $-0.377\pm 0.049$       &  {$\mathbf{-0.365 \pm 0.038}$} & $-0.368 \pm 0.039$       \\ 
 {Normalized Mean Square Error} &  {$0.055\pm 0.014$} & $0.055\pm 0.014$       &  {$\mathbf{0.054\pm 0.013}$} & ${0.054 \pm 0.013}$       \\ 
 {Running Time (s)} & {$32.39 \pm 0.35$}                &  {$32.29\pm 0.28$} & {$\mathbf{5.11 \pm 0.07}$}                & {$\mathbf{5.09\pm 0.06}$} \\ \bottomrule\end{tabular}
\label{tab:experiment_1_results}
\end{table*}

We first experiment with synthetic datasets to verify two of our main assertions: (1) that the use of linear time GPs can increase performance and (2) that the use of arithmetic averaging can increase predictive performance, in terms of predictive likelihoods and squared error of the mean. The first experiment is with outliers only, where $N_\text{data} = \num{3000}$ data points are generated as follows: first, values of $t$ are sampled from a uniform distribution $\mathcal{U}(0, 3000)$ and sorted. Next, $y$ values are sampled from a GP with a rather expressive kernel. Our focus is that the resulting functions are complex rather than the specific choice of kernel, but for completeness, a mixture Hida-Mat\'ern-3/2 kernel of order 3 \cite{dowling2021hida} is used.
All $y$ values are then given white Gaussian observation noise with variance $0.3^2$, except for \num{10} random ``true outliers,'' where the variance of the observation noise was instead $2.0^2 + 0.3^2$. We use $10$ realizations of the GP corresponding to different random seeds, pictured in \cref{fig:experiment_1_data}.

The synthetic data provide a convenient ground truth for comparison. Namely, we measure the mean predictive log-likelihood $\text{MPLL} = \frac{1}{N_\text{data}} \sum_n p(y_n | m_n, s_n^2)$, and the normalized mean square error $\text{nMSE} = \frac{1}{N_\text{data}} \sum_n (m_n - y_n)^2 / s_y^2$, where $s_y^2$ is the variance of $y$. Since the outliers are known, we exclude them in these calculations. We use two kernels for both \textsc{Intel} and \textsc{Lintel}: one is a simple Mat\'ern-5/2 kernel, which provides reasonable but suboptimal performance alone, and the other is an evidence-maximized kernel from the mixture of Hida-Mat\'ern-3/2 family. Once again, our focus is not on these specific choices, but rather, that the ``true'' kernel is available alongside a ``reasonable,'' but incorrect, kernel. Evidence maximization is performed on the first \num{250} samples, and the rest are used for online inference. An example of the resulting predictions is pictured in \cref{fig:experiment_1_example}. In this experiment, we set $L$ to be larger than the length of the time series (i.e., the mean will never periodically update), choose a maximum PCB size of $N_\text{PCB} = 3$, and set $\tau = 20$.

The results of this experiment can be found in \cref{tab:experiment_1_results}, and support both of our claims --- namely, that arithmetic fusion and the linear GP formulation outperform \textsc{Intel}, and that \textsc{Lintel} is dramatically faster than \textsc{Intel}.

\subsection{Synthetic Data With Outliers and Regime Switches} \label{sec:experiments_syn_with_regime}
We next experiment with a synthetic dataset that contains both outliers and regime switches. The dataset is quite similar to that of \cref{sec:experiments_syn_only_outliers}, except $y_{1500}, \dots, y_{2000}$ are drawn from a GP with a simple Mat\'ern-5/2 kernel, rather than the more complex Hida-Mat\'ern kernel. Since a regime switch is expected, we set $L = 250$ in this experiment.

\begin{figure*}[th]
    \centering
    \includegraphics[width=0.8\textwidth]{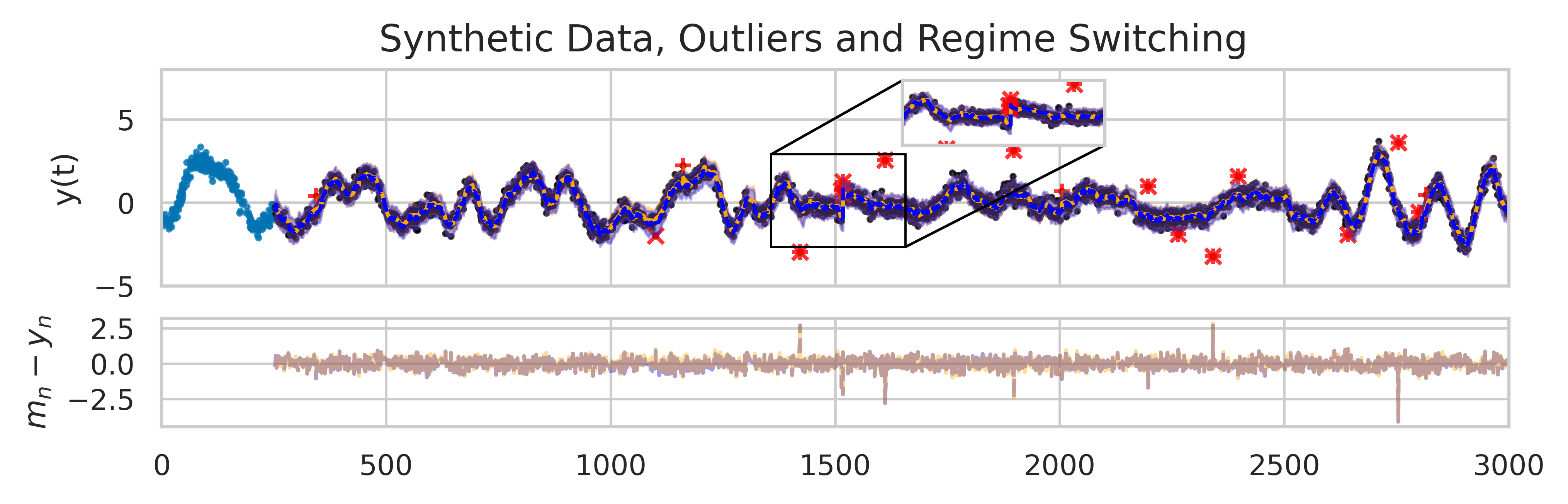}
    \caption{An example of the output for the synthetic data with outliers and regime switching experiment. {\textbf{Top}:} The outputs of \textsc{Intel} and \textsc{Lintel}, with reported outliers marked. Shaded regions denote two standard deviations. {\textbf{Bottom:}} The difference in predictive mean $m_n$ and the data point $y_n$. The legend is the same as \cref{fig:experiment_1_example} and is therefore omitted.}
    \label{fig:experiment_3_fig}
\end{figure*}

An example plot of results for one realization is pictured in \cref{fig:experiment_3_fig}. The numerical results are extremely similar to the experiment of the previous section, so we do not report them here --- the key takeaways are the same: \textsc{Lintel} achieves quantifiably higher MPLL and lower nMSE (significant at the $p=0.01$ level according to a Wilcoxon signed rank-sum test) in approximately $1/7$th the time.

We make the following remark regarding the performance of both \textsc{Intel} and \textsc{Lintel}:

\paragraph*{Remark 1} In this experiment, regime switches are only occasionally detected, as the dynamic weighting mechanism often adjusts before three consecutive outliers are observed. This may be desirable if predictive performance is the paramount metric, but undesirable if the identity of change points is of importance.

\subsection{CPU Utilization Dataset} \label{sec:experiments_real_data}

\begin{figure*}[th]
    \centering
    \includegraphics[width=0.85\textwidth]{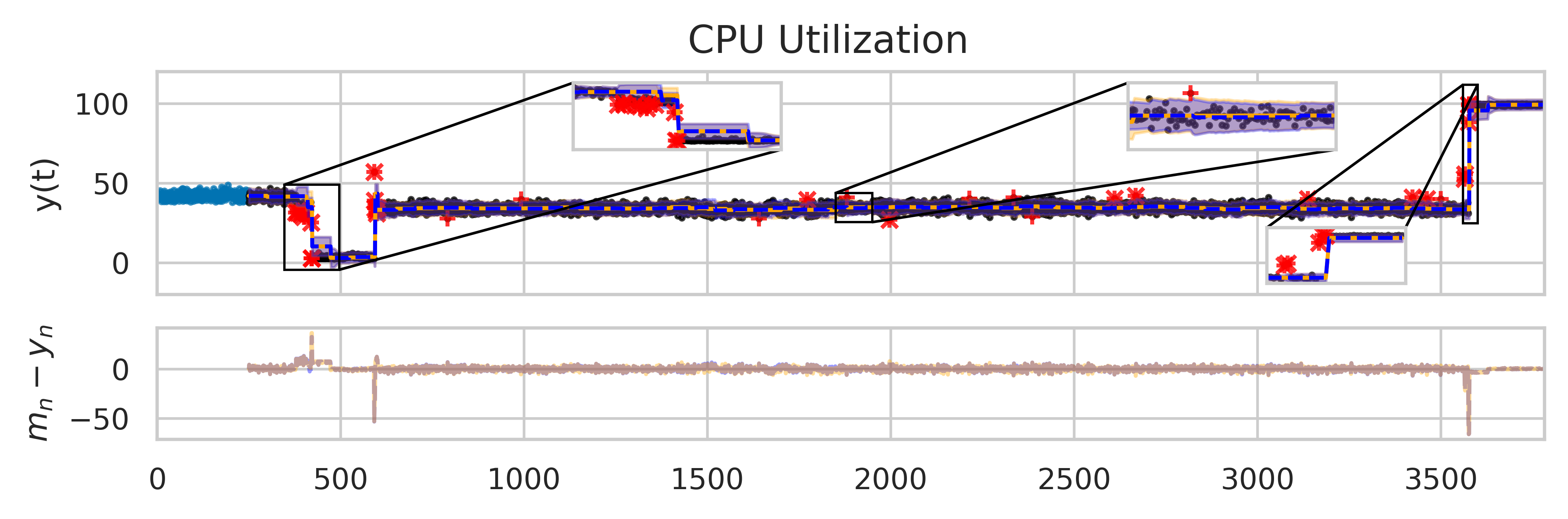}
    \caption{Output of \textsc{Intel} and \textsc{Lintel} on the CPU utilization dataset. {\textbf{Top}:} The outputs of \textsc{Intel} and \textsc{Lintel}, with reported outliers marked. Shaded regions denote two standard deviations. {\textbf{Bottom:}} The difference in predictive mean $m_n$ and the data point $y_n$. The legend is the same as \cref{fig:experiment_1_example} and is therefore omitted.}
    \label{fig:experiment_2}
\end{figure*}

We additionally test \textsc{Intel} and \textsc{Lintel} on the CPU utilization dataset, mirroring the analysis in the \textsc{Intel} paper. The CPU utilization dataset, found in the Numenta Anomaly Benchmark (NAB) dataset\footnote{Specifically, we use \texttt{ec2\_cpu\_utilization\_ac20cd}} \cite{ahmad2017unsupervised}, is a real-world dataset with a simple structure but several regime switches. For this dataset, we use candidate models similar to those in the \textsc{Intel} paper. In particular, we use a set of eight Mat\'ern-3/2 kernels, where the process variance and variance of the observation noise are set to twice or half the value of the evidence-maximized kernel. Here, we set $L = 50$ and $N_{\text{PCB}} = 3$, and use arithmetic averaging for \textsc{Lintel} and geometric averaging for \textsc{Intel}.

The results can be found in \cref{fig:experiment_2}. Overall, the results are extremely similar, with the notable exception of speed: \textsc{Lintel} takes {\bf 27 seconds} while \textsc{Intel} takes {\bf 195 seconds}. This speedup is consistent with that of the synthetic data experiments and shows significant promise in higher-throughput online inference.

Of particular note is that the change points detected by \textsc{Lintel} and \textsc{Intel} are the same, and that predictions are largely the same. This is intuitive given the nature of the function: such low-variance functions do not suffer harshly from finite context windowing.

\section{Conclusion \& Future Directions} \label{sec:conclusions}
In this paper, we introduced \textsc{Lintel}, a modified version of \textsc{Intel} with constant-time updates using Markovian GPs. We showed how we can increase computational efficiency while also circumventing approximations to the predictive distributions. The state-space formulation of \textsc{Lintel} in terms of Markovian GPs also provides connections to the robust filtering literature.

We evaluated the proposed method on two synthetic datasets, showing superior performance when the ground truth is known, and on a real-world dataset, showing nearly identical prediction at $7\times$ higher throughput. Future work may consider further benchmarking, for example, on the entire NAB dataset.

During our experiments, we occasionally noted the numerical ``collapse'' of the weights vector $\V{w}$, in the sense that a weight would numerically underflow to zero, in which case it cannot be revived, even when working with 64-bit floating point numbers. As a fix in the code, we simply add a small perturbation to the weight matrix at each time step. However, this problem was also noticed and considered in similar online ensembling problems \cite{waxman2024doebe}, and future research may incorporate the solutions proposed there. 

In this work, we showed that using BMA-derived weights was more conducive to arithmetic averaging than geometric averaging when evaluating the predictive log-likelihood. However, instead of abandoning geometric averaging, it is also interesting to consider deriving weights as an online optimization problem. While na\"ively one would consider weights that sum to unity, it has recently been noted in the literature that the generalized product of experts for GPs may benefit from allowing $\V{w}$ to not sum to unity, as this allows finer control of the predictive variance \cite{ajirak2023fusion}. It is then interesting to consider a version of \textsc{Lintel} where the product of experts fusion rule is maintained.

\bibliographystyle{IEEEtran}
\bibliography{references}

\end{document}